\documentclass[runningheads]{llncs}
\usepackage{graphicx}
\usepackage{booktabs} % For formal tables
\usepackage{amsmath,amssymb,amsfonts}
\usepackage{algorithmic}
\usepackage[linesnumbered, ruled]{algorithm2e}
\usepackage{siunitx}
\usepackage{pifont}
\usepackage{diagbox}
\usepackage{multirow}

\usepackage{float}
\usepackage{tikz}
\usetikzlibrary{positioning}
\usepackage{verbatim}

\definecolor{lightgray}{gray}{0.9}

\newcommand*{\circled}[1]{\lower.7ex\hbox{\tikz\draw (0pt, 0pt)%
		circle (.4em) node {\makebox[1em][c]{\scriptsize #1}};}}

\usepackage{cite}
\usepackage{bbding}
\usepackage{tikz,xcolor,hyperref}   
\definecolor{lime}{HTML}{A6CE39}
\DeclareRobustCommand{\orcidicon}{
	\hspace{-2.5mm}
	\begin{tikzpicture}
	\draw[lime, fill=lime] (0,0) 
	circle [radius=0.16] 
	node[white] {{\fontfamily{qag}\selectfont \tiny ID}};
	\draw[white, fill=white] (-0.0625,0.095) 
	circle [radius=0.007];
	\end{tikzpicture}
	\hspace{-2mm}
}

\foreach \x in {A, ..., Z}{%
	\expandafter\xdef\csname orcid\x\endcsname{\noexpand\href{https://orcid.org/\csname orcidauthor\x\endcsname}{\noexpand\orcidicon}}
}

\begin{document}
%
% \title{Contribution Title\thanks{Supported by organization x.}}
\title{Investigation of the Generalisation Ability of Genetic Programming-evolved Scheduling Rules in Dynamic Flexible Job Shop Scheduling}
% %
\titlerunning{Investigation of the Generalisation of GP-evolved Scheduling Rules in DFJSS}
%\title{An Effective Sampling Strategy for Surrogate-assisted Genetic Programming for Dynamic Flexible Job Shop Scheduling}
%\titlerunning{An Effective Sampling Strategy for Surrogate assisted GP for DFJSS}

%\author{Fangfang Zhang$^{1}\textsuperscript{\Envelope}$\orcidA{} \and Yi Mei$^{1}$\orcidB{} \and Su Nguyen$^{2}$\orcidC{} \and Mengjie Zhang$^{1}$\orcidD{}}

\author{
Luyao Zhu \and
Fangfang Zhang \and 
Yi Mei \and
Mengjie Zhang}

\authorrunning{L. Zhu et al.}

\institute{ Centre for Data Science and Artificial Intelligence \&
School of Engineering and Computer Science, Victoria University of Wellington, Wellington 6140, New Zealand\\}

%\author{Luyao Zhu$^{1}$ \and Fangfang Zhang$^{2}$\orcidA{}  \and Xiaodong Zhu$^{1}$ \and Ke Chen$^{1}$\textsuperscript{\Envelope}}

%\authorrunning{L. Zhu et al.}

%\institute{
%$^{1}$ School of Electrical Engineering, Zhengzhou University, Zhengzhou 450001, China \\
%\email{zhuluyao58@163.com,\{Zhu\_xd,chenkezixf\}@zzu.edu.cn}\\
%$^{2}$ School of Engineering and Computer Science, Victoria University of Wellington, \\
%PO BOX 600, Wellington 6140, New Zealand \\
%\email{fangfang.zhang@ecs.vuw.ac.nz} \\
%} 

\maketitle              % typeset the header of the contribution
\begin{abstract}
% The abstract should briefly summarize the contents of the paper in
% 15--250 words.
Dynamic Flexible Job Shop Scheduling (DFJSS) is a complex combinatorial optimisation problem that requires simultaneous machine assignment and operation sequencing decisions in dynamic production environments. Genetic Programming (GP) has been widely applied to automatically evolve scheduling rules for DFJSS. However, existing studies typically train and test GP-evolved rules on DFJSS instances of the same type, which differ only by random seeds rather than by structural characteristics, leaving their cross-type generalisation ability largely unexplored. To address this gap, this paper systematically investigates the generalisation ability of GP-evolved scheduling rules under diverse DFJSS conditions. A series of experiments are conducted across multiple dimensions, including problem scale (i.e., the number of machines and jobs), key job shop parameters (e.g., utilisation level), and data distributions, to analyse how these factors influence GP performance on unseen instance types. The results show that good generalisation occurs when the training instances contain more jobs than the test instances while keeping the number of machines fixed, and when both training and test instances have similar scales or job shop parameters. Further analysis reveals that the number and distribution of decision points in DFJSS instances play a crucial role in explaining these performance differences. Similar decision point distributions lead to better generalisation, whereas significant discrepancies result in a marked degradation of performance. Overall, this study provides new insights into the generalisation ability of GP in DFJSS and highlights the necessity of evolving more generalisable GP rules capable of handling heterogeneous DFJSS instances effectively.

\end{abstract}

\section{Introduction}
Dynamic Flexible Job Shop Scheduling (DFJSS) \cite{shen2015mathematical} is an important combinatorial optimisation problem that aims to process multiple jobs using a set of machines. In DFJSS, each job consists of multiple operations, and each operation can be processed on more than one machine. Therefore, two key decisions, i.e., machine assignment and operation sequencing, must be made simultaneously. Moreover, these decisions are often required in dynamic environments \cite{yi2025dynamic}, where jobs or batches arrive over time. This study focuses on heterogeneous batch arrivals \cite{zhu2025batch}, where diverse jobs dynamically arrive in batches with varying characteristics. Genetic Programming (GP) \cite{koza1990genetic,koza2005genetic} has been widely applied to automatically evolve scheduling rules for DFJSS \cite{DBLP:journals/gpem/DurasevicJ18,10065588}. Specifically, in DFJSS, GP is used to evolve a pair of scheduling rules: a routing rule for machine assignment and a sequencing rule for operation sequencing.

In DFJSS, GP-based learning typically involves two stages: training and test. During training, GP evolves scheduling rules, with each individual representing a pair of scheduling rules. These individuals are evaluated on a set of training instances (i.e., DFJSS simulations) \cite{hildebrandt2010towards} to measure their performance. The best-evolved scheduling rule pair is then selected. In the test stage, the best rule pair is applied to unseen test instances to get the scheduling solutions.

However, most existing studies use training and test instances with identical settings, including the same problem scale (e.g., number of machines and jobs), job shop parameters (e.g., utilisation level and due date factor), and data distributions (e.g., operation processing times drawn from a uniform distribution between 1 and 99). These instances differ only in random seeds. As a result, the generalisation ability reported in these studies is restricted to same-type generalisation, providing limited evidence on whether the evolved rules can perform effectively across different types of instances. Several studies have explored the generalisation ability of GP-evolved rules under different conditions. Mei et al. \cite{mei2016comprehensive} conducted a comprehensive experimental analysis to examine the reusability of GP-evolved rules on unseen instances with varying numbers of jobs and machines, but their work was limited to static and small-scale job shop scheduling (JSS) problems. Park et al. \cite{park2016investigating} investigated the generalisation ability of GP in Dynamic JSS under different machine breakdown ratios and found that GP often fails to generalise effectively in highly dynamic environments. They also observed that the performance of standard GP-based hyper-heuristics tends to be sensitive to machine unavailability during simulations.

However, there has been no comprehensive investigation into the generalisation ability of GP in DFJSS under diverse conditions. To fill this gap, this paper conducts extensive experimental studies to systematically examine this issue and aims to answer the following research questions.

\begin{itemize}
\item How does \textit{problem scale} affect the generalisation ability of GP-evolved rules?
\item How does \textit{workshop parameter} influence the generalisation ability of GP-evolved rules?
\item How does \textit{parameter distribution} impact the generalisation ability of GP-evolved rules?
\item What underlying factors contribute to the observed performance differences, and what insights can be drawn from the results?
\end{itemize}

\section{Background}

\subsection{Problem Statement}

In a DFJSS environment, a set of machines 
$\mathcal{M} = \{M_1, M_2, ..., M_m\}$ is responsible for processing batches of jobs 
$\mathcal{B} = \{B_1, B_2, ..., B_n\}$. 
Each batch $B_i$ consists of multiple jobs 
$\mathcal{J}_i = \{J_{i1}, J_{i2}, ..., J_{ij}\}$, 
which arrive dynamically at the shop floor in a batch-by-batch manner. 
All jobs within a batch are independent and may correspond to different customer orders. 
The details of each batch are unknown until its arrival on the shop floor. Each job $J_{ij}$ comprises a predefined sequence of operations 
$\mathcal{O}_{ij} = \{O_{ij1}, O_{ij2}, ..., O_{ijr}\}$. 
In DFJSS, each operation $O_{ijr}$ can be processed by multiple machines, 
represented by the set $M(O_{ijr}) \subseteq \mathcal{M}$ \cite{brucker1990job}.

The optimisation objective of this study is to minimise the 
\textit{mean-weighted-tardiness} (WT$_{\text{mean}}$) of all jobs, which is formulated as:

\begin{equation}
WT_{\text{mean}} = \frac{1}{|J|} \sum_{J_{ij} \in J} w_{ij} \times \max \{0,\, C_{ij} - d_{ij} \}
\end{equation}

\noindent
where $J$ denotes the set of all jobs $J_{ij}$, $w_{ij}$ is the weight of job $J_{ij}$, $C_{ij}$ represents its completion time, and $d_{ij}$ is its due date. In addition, the following main constraints are considered in the DFJSS problem:

\begin{itemize}
    \item An operation cannot start before the completion of its preceding operation.  
    \item Each operation must be assigned to one machine from its eligible set. 
    \item Each machine can process at most one operation at any given time.  
    \item Once an operation starts on a machine, it cannot be stopped.
\end{itemize}

\subsection{GP for DFJSS}

\begin{figure}[t]
	\centering 
	\includegraphics[width= 1\textwidth]{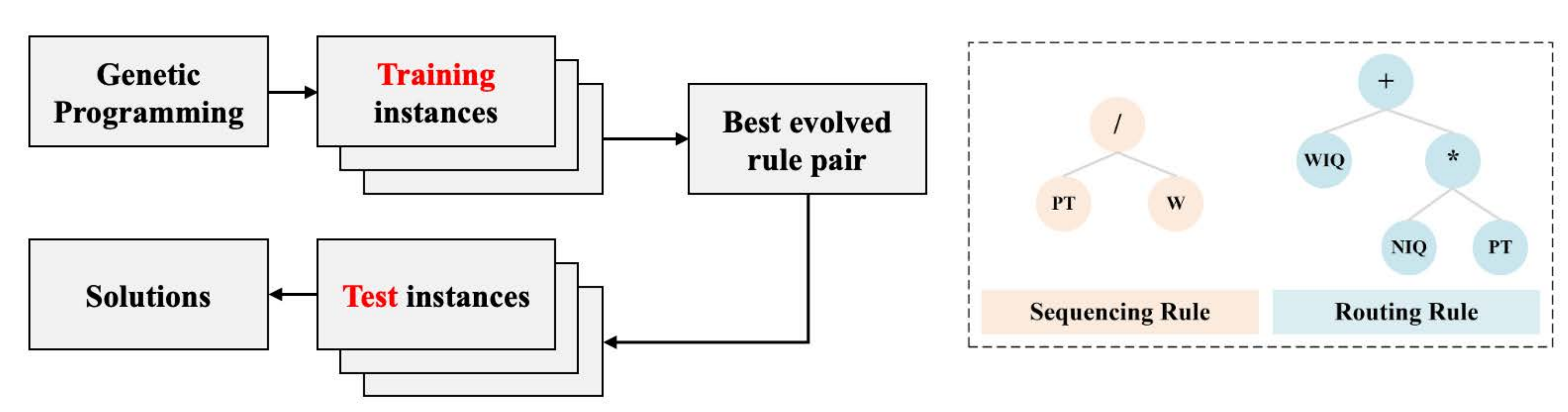}
	\vspace{-3mm}
	\caption{The whole process of GP for DFJSS and an example of a GP individual.} \label{Fig. TrainTest}
\end{figure}

GP has been successfully applied as a hyper-heuristic approach to learn scheduling heuristics for DFJSS problems \cite{zhang2020FS,10612212,DBLP:journals/asc/DurasevicJK16}. The GP process for DFJSS consists of two stages, training and test, as illustrated in Fig.~\ref{Fig. TrainTest}. The training process includes initialisation, evaluation, parent selection, and offspring generation. During evaluation, GP-evolved rules are applied to DFJSS training instances to measure their performance. Once the stopping criterion is met, GP outputs the best scheduling rule pair found. After that, the best scheduling rule pair is applied to DFJSS test instances to generate the final solutions.

The multi-tree representation is adopted to evolve scheduling rules for DFJSS \cite{zhang2018genetic}. Each GP individual contains two trees (an example is presented in the right of Fig.~\ref{Fig. TrainTest}): one for sequencing and one for routing. The individual’s fitness is evaluated based on their interaction. The sequencing rule example is $PT/W$, where $PT$ is the processing time of an operation on a machine and $W$ is the job weight. The routing rule example is $WIQ + (NIQ \times PT)$, where $WIQ$ and $NIQ$ denote the workload and the number of operations in a machine’s queue. These rules assign priorities to candidate operations or machines, and the one with the lowest value is selected~\cite{11043103}.

\subsection{Related Work about Generalisation of GP-evolved Rules in JSS}

There are several studies that focus on analysing the generalisation ability of GP-evolved rules in JSS problems. Nguyen et al.~\cite{nguyen2013computational} applied GP-evolved rules obtained from static JSS instances to dynamic JSS problems with random job arrivals and showed that they did not perform as well as manually designed dispatching rules. This indicates that GP-evolved rules trained on static JSS problems cannot generalise well to dynamic JSS problems. Mei et al.~\cite{mei2016comprehensive} conducted the first systematic investigation into the generalisation ability of GP-evolved scheduling rules. They performed a comprehensive experimental analysis to examine the generalisation ability of GP-evolved rules on unseen test instances with varying numbers of jobs and machines. Their results demonstrated that better reusability can be achieved by selecting training instances whose numbers of jobs and machines (or at least the ratio between them) are closer to those of the test instances. In addition, Nguyen et al. \cite{nguyen2017surrogate} investigated dynamic JSS instances with proportionally smaller numbers of jobs and machines, showing that such surrogate instances can also achieve good generalisation. Park et al.~\cite{park2016investigating} examined the generalisation ability of GP in dynamic JSS problems under different machine breakdown ratios and concluded that GP may fail to generalise effectively in scenarios with dynamic arrivals and machine breakdowns. They also observed that the performance of rules evolved by GP tends to be sensitive to the proportion of time that machines remain unavailable during simulations. 

However, the above studies are limited to limited conditions. To address this, this paper systematically investigates the generalisation ability of GP-evolved rules in DFJSS by training on one instance type and testing on others across varying problem scales, problem parameters, and parameter distributions.

\section{Experiment Design}
\subsection{Simulation Model}

For DFJSS, a simulation-based approach~\cite{kiran1984simulation} is used to model the system, where each run represents one instance. In the simulation, $m$ machines process $n$ jobs, and their numbers define the problem scale. Jobs arrive in batches, with batch sizes following a distribution over $[b_i, b_j]$~\cite{zhu2025batch}. To eliminate initialisation effects, the first one-sixth of batches are treated as a warm-up phase, and the remaining batches are used for evaluation. Fixed dispatching rules, Shortest Processing Time (SPT) for sequencing and Lowest Workload in Queue (LWIQ) for routing, are applied to ensure consistent conditions after the warm-up phase.

Batch arrivals follow a Poisson process, meaning that inter-arrival times are exponentially distributed. This stochastic assumption is commonly used in job shop simulations to capture job arrival randomness. The mean inter-arrival time is adjusted to achieve the target system utilisation $u$. Job importance is represented by weights: 20\% of jobs have a weight of 1 (low priority), 60\% a weight of 2 (medium), and 20\% a weight of 4 (high)~\cite{DBLP:journals/ec/HildebrandtB15}. The number of candidate machines per operation is drawn from $[1, m]$, and the number of operations per job from $[1, m]$. Processing times are independently sampled from $[1, 99]$ and vary across machines by roughly 10 time units. Each job’s due date equals its estimated processing time multiplied by a due-date factor $dd$, plus its arrival time. Specific parameter values are detailed in the following section.

\subsection{Instance Design}

To investigate the generalisation ability of GP-evolved rules across different DFJSS problems, three types of training and test instance settings are designed: problem scale, problem parameters, and parameter distributions. 

\begin{itemize}
    \item \textbf{Problem Scale:} This setting examines how GP-evolved rules generalise across different problem scales. The number of jobs and machines is varied to create instances of different scales, while all other factors are kept constant.

    \item \textbf{Problem Parameters:} This setting investigates the effect of varying specific job shop characteristics. Key parameters include the utilisation level (representing the workload intensity of the system), the due date factor (indicating delivery urgency), and the batch size (reflecting the level of instantaneous congestion). All remaining conditions are held constant.

    \item \textbf{Parameter Distributions:} Considering the stochastic nature of job shop environments, this setting explores the influence of different sampling distributions. Five types of probability distributions (e.g., uniform, normal, exponential) are employed to generate specific data.
\end{itemize}

\subsection{GP Parameter Settings}

Each GP individual comprises terminals and functions. The terminals represent job shop features related to machines, operations, and jobs/batches, consistent with those used in~\cite{zhu2025batch}. The function set is defined as \{\(+\), \(-\), \(*\), protected \(/\), \(max\), \(min\)\}. The protected division returns one when the denominator is zero. Other GP parameter settings follow the recommendations in \cite{zhu2025batch}, the population size and number of generations are 500 and 100, separately. We use the ramp-half-and-half method to generate initialised GP programs with a minimal (maximal) depth of 2 (6). The maximal depths of all GP individuals are 8. Tournament selection with size 7 is used to select parents. Ten elites will be preserved to the next generation, the other new offspring are generated by genetic operators, i.e. crossover, mutation and reproduction with rates 80\%, 15\%, and 5\%, respectively.

%as detailed in Table~\ref{tab:terminals}.
% \begin{table} [t]
%     \centering 
% 	\caption{The terminal set.}	\label{tab:terminals} 
% %	\vspace{-3mm}
% 	\footnotesize
% 	\begin{tabular}  {c c}
% 		\toprule 
% 		\bfseries Notation& \bfseries Description\\
% 		\midrule 
		
% 		MWT &  A machine's waiting time\\
%         WIQ & Current work in the queue\\
%         NIQ  &  The number of operations in the queue \\ 
%         NPT & Median processing time for the next operation \\
%         OWT & The waiting time of an operation \\
        
% 		PT  & Processing time of an operation on a specified machine \\
% 		WKR & Median amount of work remaining for a job \\
%             NOR & The number of operations remaining for a job \\
% 		TIS &Time in system\\
%         W & Weight of a job \\
%         rDD & Relative due date of a job \\
% 		  SL  &  Slack of a job \\
%         BNOR & The number of operations remaining of a batch \\
%         BWKR &  Median amount of work remaining of a batch \\
%         NCM & The number of candidate machines for an operation\\
		
% 		\bottomrule
% 	\end{tabular}
% \end{table}

\section{Results and Analyses}
Due to the stochastic nature of evolutionary algorithms, each experiment is repeated 30 times independently to ensure statistical reliability. The Wilcoxon rank-sum test with a 0.05 significance level is used to assess performance differences. In the following results, the symbols ``$\uparrow$'', ``$\downarrow$'', and ``$\approx$'' denote results that are significantly better, worse, or similar to the \textbf{reference result in each row}, respectively. The reference corresponds to the case where the training and test instances are identical and is highlighted in grey in the subsequent tables.

\subsection{Different Problem Scale}

In this part, the training and test instances differ in the number of jobs and machines. Specifically, three types of scale variations are considered:  

\begin{itemize}
\item Different numbers of jobs with a fixed number of machines.
\item Different numbers of machines with a fixed number of jobs.
\item Different numbers of jobs and machines while keeping their ratio constant.
\end{itemize}

\subsubsection{Varying the Number of Jobs with a Fixed Number of Machines}

\label{job}

Ten machines are used to process five job sizes: 100, 1000, 2500, 5000, and 10000. Table~\ref{tab:job} presents the test performance across different job numbers. The notation $\langle n, m \rangle$ represents a DFJSS instance with $n$ machines and $m$ jobs. Each column corresponds to a model trained on one instance type and tested on all five. For example, the column $\langle 10, 100 \rangle$ indicates that GP is trained on $\langle 10, 100 \rangle$ and tested on $\langle 10, 100 \rangle$, $\langle 10, 1000 \rangle$, $\langle 10, 2500 \rangle$, $\langle 10, 5000 \rangle$, and $\langle 10, 10000 \rangle$.

\begin{table}[t] 
	\centering
	\footnotesize 
% 	\tiny
% 	\scriptsize
	%\setlength{\tabcolsep}{3mm}
	\caption{The mean (standard deviation) of objective values on test instances according to 30 independent runs \textbf{across different number of jobs}.}
	
	\vspace{-3mm}	
	\label{tab:job} 
	%\scalebox{1}
	\resizebox{\textwidth}{!}{
		\newcommand{\tabincell}[2]{\begin{tabular}{@{}#1@{}}#2\end{tabular}}
		\begin{tabular}  { cccccc} 
			\toprule 
			\bfseries 	\footnotesize   
            \diagbox[width=4em]{Test}{Train} &
              \bfseries 	\footnotesize  $\langle$10, 100$\rangle$ &
              \bfseries   \footnotesize  $\langle$10, 1000$\rangle$ & 
              \bfseries 	\footnotesize  $\langle$10, 2500$\rangle$  & 
              \bfseries 	\footnotesize  $\langle$10, 5000$\rangle$ & 
              \bfseries 	\footnotesize  $\langle$10, 10000$\rangle$ \\
			\midrule
            
	\bfseries $\langle$10, 100$\rangle$ & \colorbox{lightgray}{297.64(14.54)}& 286.00(9.95) {($\uparrow$)}  & 288.62(13.25) {($\uparrow$)} & 279.98(8.87) {($\uparrow$)} & 282.65(11.44) {($\uparrow$)} \\
    
    \bfseries $\langle$10, 1000$\rangle$ &474.29(45.27) {\bf($\downarrow$)} & 
	\colorbox{lightgray}{430.83(28.27)} & 436.60(33.11){($\approx$)}& 
	419.63(21.97){\bf($\uparrow$)}&425.40(30.28) {($\uparrow$)} \\
    
	\bfseries $\langle$10, 2500$\rangle$ &443.27(40.67) {\bf($\downarrow$)} & 
	402.90(25.26){($\approx$)} & \colorbox{lightgray}{408.58(30.13)} & 
	392.75(19.29){\bf($\uparrow$)} & 397.98(26.94) {\bf($\uparrow$)} \\
    
	  \bfseries $\langle$10, 5000$\rangle$ & 429.46(38.99){($\downarrow$)} & 
	390.79(24.33){($\downarrow$)} & 396.23(28.99){($\downarrow$)}& 
	\colorbox{lightgray}{380.87(18.68)} & 386.05(25.98) {($\approx$)} \\
	\bfseries $\langle$10, 10000$\rangle$& 412.67(36.83) {($\downarrow$)} & 
	376.27(23.15){($\downarrow$)} & 381.52(27.74){($\approx$)}& 
	366.96(17.79){($\approx$)}&\colorbox{lightgray}{371.86(24.82)}\\
	
	\bottomrule 
	\end{tabular}}
\end{table}

From the results, an interesting pattern emerges. When the training instances contain more jobs than the test instances, the test performance tends to surpass that of the same-scale instances (highlighted in grey). This is evident from the rows where results to the right of the grey cells are generally marked with ($\uparrow$). In contrast, when the training instances contain fewer jobs than the test instances, performance is usually worse, as shown by the ($\downarrow$) marks to the left of the grey cells. Additionally, several ($\approx$) symbols appear in cases with neighbouring job numbers, indicating comparable performance.

This pattern can be explained by the number of decision points during scheduling. Increasing the number of jobs does not alter the overall decision point distributions (e.g., remaining workload or unfinished operations) but naturally generates more decision points. Consequently, GP trained on larger instances encounters a richer and more diverse set of scheduling situations, enabling it to learn more general decision-making patterns and achieve better generalisation across scales. In contrast, training on smaller instances offers fewer decision points and limited diversity, leading to overfitting to simpler dynamics and poorer performance on larger instances. %When the training and test instances have neighbouring scales （e.g., 1000 and 2500, 5000 and 10000）, the number of decision points does not differ too much, producing similar scheduling contexts and thus comparable performance.
 
\subsubsection{Varying the Number of Machines with a Fixed Number of Jobs}
\label{machine}
In this section, the number of jobs is fixed while the number of machines varies. The maximum number of operations per job is set equal to the number of machines. Five configurations are considered, with 2, 5, 10, 20, and 50 machines used to process 5,000 jobs. Table~\ref{tab:machine} presents the test performance across different machine numbers. Unlike the job-scale experiments, where rules trained on larger instances generalised well to smaller ones, this pattern is not observed here. The results show that the best performance is typically achieved when the training and test instances have the same number of machines. When the number of machines differs, either larger or smaller, performance declines significantly.
\begin{table}[t] 
	\centering
	\footnotesize 
	\caption{The mean (standard deviation) of objective values on test instances according to 30 independent runs \textbf{across different number of machines}.}
	
	\vspace{-3mm}	
	\label{tab:machine} 
	\resizebox{\textwidth}{!}{
		\newcommand{\tabincell}[2]{\begin{tabular}{@{}#1@{}}#2\end{tabular}}
		\begin{tabular} { cccccc} 
			\toprule 
			\bfseries \footnotesize 	
			\diagbox[width=4em]{Test}{Train} &
			\bfseries 	\footnotesize $\langle$2, 5000$\rangle$ &
			\bfseries 	\footnotesize $\langle$5, 5000$\rangle$ &	
			\bfseries 	\footnotesize $\langle$10, 5000$\rangle$	 &	
			\bfseries 	\footnotesize $\langle$20, 5000$\rangle$ &	
			\bfseries 	\footnotesize $\langle$50, 5000$\rangle$ \\
			\midrule
			
			% --- 第 1 行：Test <2, 5000> ---
			\bfseries $\langle$2, 5000$\rangle$ & 
			\colorbox{lightgray}{638.61(2.02)} & 
			653.68(31.37) {($\downarrow$)} & 
			689.80(66.47) {($\downarrow$)} & 
			742.50(92.73) {($\downarrow$)} & 
			1003.12(250.97) {($\downarrow$)} \\
			
			% --- 第 2 行：Test <5, 5000> ---
			\bfseries $\langle$5, 5000$\rangle$ & 
			526.70(13.24) {($\downarrow$)} &	
			\colorbox{lightgray}{483.38(21.84)} & 
			488.89(25.65){($\approx$)}&	
			534.36(51.52){($\downarrow$)}&
			708.47(137.65) {($\downarrow$)} \\
			
			% --- 第 3 行：Test <10, 5000> ---
			\bfseries $\langle$10, 5000$\rangle$ &
			550.21(42.43) {($\downarrow$)} &	
			414.00(40.08) {($\downarrow$)} & 
			\colorbox{lightgray}{380.88(18.69)} &	
			408.58(34.80){($\downarrow$)} & 
			513.92(85.93) {($\downarrow$)} \\
			
			% --- 第 4 行：Test <20, 5000> ---
			\bfseries $\langle$20, 5000$\rangle$& 
			668.58(231.51){($\downarrow$)} &	
			331.41(82.84){($\downarrow$)} & 
			245.06(15.25){($\approx$)}&	
			\colorbox{lightgray}{249.52(21.03)} & 
			287.05(34.59) {($\downarrow$)} \\
			
			% --- 第 5 行：Test <50, 5000> ---
			\bfseries $\langle$50, 5000$\rangle$& 
			1036.23(575.52) {($\downarrow$)} &	
			219.21(158.51){($\downarrow$)} & 
			73.39(20.10){($\downarrow$)}&	
			64.50(8.42){($\approx$)}&
			\colorbox{lightgray}{61.01(6.91)}\\
			
			\bottomrule 
		\end{tabular}
	}
\end{table}

Although more machines increase the total number of decision points, their distribution changes substantially. With fewer machines, machines are busier and competition among operations intensifies. In contrast, more machines reduce the workload per machine and alter the scheduling dynamics. In addition, the candidate-machine sets for operations differ across configurations, causing a shift in decision point distributions. As a result, the scheduling situations encountered during training are not representative of those during testing, and GP-evolved rules trained under one machine configuration fail to generalise well to others.

\subsubsection{Varying the Number of Machines and Jobs with a Fixed Ratio}
\label{ratio}

According to the conclusion in \cite{mei2016comprehensive}, better reusability of GP-evolved rules can be achieved by selecting training instances whose ratio between machines and jobs is closer to that of the test instances. However, their experiments were conducted in static and small-scale JSS problems. It remains unclear whether this conclusion still holds in dynamic and large-scale JSS problems. To this end, we fixed the ratio between machines and jobs to investigate this factor. Specifically, we fixed the ratio to 1:500 \cite{zhang2020FS}. Five settings were tested, i.e., $\langle 2, 1000 \rangle$, $\langle 5, 2500 \rangle$, $\langle 10, 5000 \rangle$, $\langle 15, 7500 \rangle$, and $\langle 20, 10000 \rangle$. 

Table~\ref{tab:ratio} presents the test performance across different numbers of machines and jobs while maintaining a fixed ratio. We find the same pattern as in Table~\ref{tab:machine}, where the best performance in each row is typically achieved when the training and test instances have the same scale. When the instance sizes differ, the generalisation performance drops noticeably. This indicates that even under a fixed ratio between machines and jobs, the distribution of decision situations still varies across different scales. Larger instances involve more decision points and longer dynamic horizons, leading to different characteristics in the system state distribution. Consequently, the evolved rules trained on one scale cannot easily adapt to another, despite having the same machine-to-job ratio.

\begin{table}[t] 
	\centering
	\footnotesize 
	\caption{The mean (standard deviation) of objective values on test instances according to 30 independent runs \textbf{across different numbers of machines and jobs while fixing the ratio}.}
	
	\vspace{-3mm}	
	\label{tab:ratio} 
	\resizebox{\textwidth}{!}{
		\newcommand{\tabincell}[2]{\begin{tabular}{@{}#1@{}}#2\end{tabular}}
		\begin{tabular} { cccccc} 
			\toprule 
			% --- 表头行（已加粗） ---
			\bfseries \footnotesize 	
			\diagbox[width=4em]{Test}{Train} &
			\bfseries \footnotesize $\langle$2, 1000$\rangle$ &
			\bfseries \footnotesize $\langle$5, 2500$\rangle$ &	
			\bfseries \footnotesize $\langle$10, 5000$\rangle$	 &	
			\bfseries \footnotesize $\langle$15, 7500$\rangle$ &	
			\bfseries \footnotesize $\langle$20, 10000$\rangle$ \\
			\midrule
			
			% --- 第 1 行：Test <2, 1000> ---
			\bfseries $\langle$2, 1000$\rangle$ & 
			\colorbox{lightgray}{600.83(5.29)} & 
			633.63(57.97) {($\downarrow$)} & % (-) -> \downarrow
			638.89(47.09) {($\downarrow$)} & % (-) -> \downarrow
			674.73(62.61) {($\downarrow$)} & % (-) -> \downarrow
			696.17(72.44) {($\downarrow$)} \\ % (-) -> \downarrow
			
			% --- 第 2 行：Test <5, 2500> ---
			\bfseries $\langle$5, 2500$\rangle$ & 
			508.55(20.42) {($\downarrow$)} &	% (+) -> \uparrow
			\colorbox{lightgray}{474.02(27.50)} & 
			468.68(23.11){($\approx$)}&	  % (=) -> \approx
			493.50(33.71){($\downarrow$)}&  % (-) -> \downarrow
			512.32(45.25) {($\downarrow$)} \\ % (-) -> \downarrow
			
			% --- 第 3 行：Test <10, 5000> ---
			\bfseries $\langle$10, 5000$\rangle$ &
			551.65(75.19) {($\downarrow$)} &	% (-) -> \downarrow
			421.15(40.00){($\downarrow$)} &   % (+) -> \uparrow
			\colorbox{lightgray}{380.88(18.69)} &	
			396.67(28.38){($\downarrow$)} & % (-) -> \downarrow
			407.60(31.91) {($\downarrow$)} \\ % (-) -> \downarrow
			
			% --- 第 4 行：Test <15, 7500> ---
			\bfseries $\langle$15, 7500$\rangle$& 
			566.98(143.87){($\downarrow$)} &	% (-) -> \downarrow
			353.38(59.94){($\downarrow$)} &   % (+) -> \uparrow
			282.38(14.57){($\approx$)}&	  % (+) -> \uparrow
			\colorbox{lightgray}{287.75(21.46)} & 
			292.24(22.46) {($\approx$)} \\   % (=) -> \approx
			
			% --- 第 5 行：Test <20, 10000> ---
			\bfseries $\langle$20, 10000$\rangle$& 
			633.78(218.34) {($\downarrow$)} &	% (-) -> \downarrow
			341.26(78.19){($\downarrow$)} &   % (+) -> \uparrow
			244.30(15.42){($\approx$)}&	  % (+) -> \uparrow
			244.55(19.08){($\approx$)}&    % (=) -> \approx
			\colorbox{lightgray}{246.70(20.10)}\\
			
			\bottomrule 
		\end{tabular}
	}
\end{table}

\subsection{Different Parameter Values}

In this part, the training and test instances differ in several job shop parameters, which is the most common approach for designing experimental scenarios in related studies~\cite{zhu2025batch,mohan2019review}. Specifically, three parameters are considered:

\begin{itemize} 
\item \textbf{Utilisation Level:} controls the inter-arrival time between consecutive jobs or batches. Higher utilisation implies shorter inter-arrival times, heavier workload, and increased machine congestion.

\item \textbf{Due Date Factor:} determines the tightness of job due dates. A larger value indicates more relaxed due dates and lower tardiness pressure, while a smaller one leads to tighter delivery constraints and more challenging scheduling.

\item \textbf{Batch Size:} specifies the number of jobs per batch. Larger batch sizes cause more jobs to arrive simultaneously, expanding the decision space and increasing scheduling complexity.

\end{itemize}

\subsubsection{Utilisation Level}
\label{utilisation level}

From idle to busy shop floors, five utilisation levels are set, specifically 0.50, 0.75, 0.85, 0.95, and 0.99, to investigate the generalisation ability of GP-evolved rules under different workload intensities. The results in Table~\ref{tab:utilisation} show a clear pattern: when the training and test utilisation levels are the same, the performance is generally the best (highlighted in grey). However, when the difference between training and test utilisation levels increases, the performance tends to be worse.

\begin{table}[t] 
	\centering
	\footnotesize 
	\caption{The mean (standard deviation) of objective values on test instances according to 30 independent runs \textbf{across different utilisation levels}.}
	
	\vspace{-3mm}	
	\label{tab:utilisation} 
	\resizebox{\textwidth}{!}{
		\newcommand{\tabincell}[2]{\begin{tabular}{@{}#1@{}}#2\end{tabular}}
		\begin{tabular} { cccccc} 
			\toprule 
			% --- 表头行（已加粗，使用缺失率） ---
			\bfseries \footnotesize 	
			\diagbox[width=4em]{Test}{Train} &
			\bfseries \footnotesize $\langle$0.50$\rangle$ &
			\bfseries \footnotesize $\langle$0.75$\rangle$ &	
			\bfseries \footnotesize $\langle$0.85$\rangle$	 &	
			\bfseries \footnotesize $\langle$0.95$\rangle$ &	
			\bfseries \footnotesize $\langle$0.99$\rangle$ \\
			\midrule
			
			% --- 第 1 行：Test <0.50> ---
			\bfseries $\langle$0.50$\rangle$ & 
			\colorbox{lightgray}{79.83(4.36)} & 
			79.31(4.73) {($\approx$)} &   % (=) -> \approx
			79.30(5.46) {($\approx$)} &   % (=) -> \approx
			105.50(32.90) {($\downarrow$)} & % (-) -> \downarrow
			107.88(30.44) {($\downarrow$)} \\ % (-) -> \downarrow
			
			% --- 第 2 行：Test <0.75> ---
			\bfseries $\langle$0.75$\rangle$ & 
			258.64(19.70) {($\downarrow$)} &	 % (-) -> \downarrow
			\colorbox{lightgray}{247.83(17.47)} & 
			239.31(10.37){($\approx$)}&	   % (+) -> \uparrow
			274.92(47.75){($\downarrow$)}&  % (-) -> \downarrow
			277.38(41.60) {($\downarrow$)} \\ % (-) -> \downarrow
			
			% --- 第 3 行：Test <0.85> ---
			\bfseries $\langle$0.85$\rangle$ &
			428.57(40.10) {($\downarrow$)} &	 % (-) -> \downarrow
			401.04(33.16){($\downarrow$)} &   % (+) -> \uparrow
			\colorbox{lightgray}{380.88(18.69)} &	
			416.68(54.42){($\downarrow$)} & % (-) -> \downarrow
			417.51(45.52) {($\downarrow$)} \\ % (-) -> \downarrow
			
			% --- 第 4 行：Test <0.95> ---
			\bfseries $\langle$0.95$\rangle$& 
			827.45(109.27){($\downarrow$)} &	% (-) -> \downarrow
			734.05(77.29){($\downarrow$)} &   % (+) -> \uparrow
			677.02(43.87){($\approx$)}&	  % (+) -> \uparrow
			\colorbox{lightgray}{699.99(68.33)} & 
			694.75(54.51) {($\approx$)} \\   % (=) -> \approx
			
			% --- 第 5 行：Test <0.99> ---
			\bfseries $\langle$0.99$\rangle$& 
			1211.00(204.06) {($\downarrow$)} &	% (-) -> \downarrow
			1027.78(126.84){($\downarrow$)} &   % (+) -> \uparrow
			929.40(73.86){($\approx$)}&	  % (+) -> \uparrow
			933.16(84.48){($\approx$)}&    % (=) -> \approx
			\colorbox{lightgray}{921.65(67.81)}\\
			
			\bottomrule 

		\end{tabular}
	}
\end{table}

In particular, the rules trained under low utilisation levels (e.g., 0.50 and 0.75) perform poorly when tested in high-utilisation environments (e.g., 0.85, 0.95 and 0.99). This is because, under low utilisation, the shop floor is relatively idle, resulting in fewer conflicts among machines and less tight scheduling pressure. The GP-evolved rules under such relaxed conditions may overfit to easy situations and fail to handle the high congestion and competition in busy environments. Conversely, rules evolved in high-utilisation settings also do not generalise well to low-utilisation environments. When the shop floor is busy, GP tends to evolve strategies that aggressively minimise waiting time or prioritise bottleneck machines. These strategies may become unnecessarily complex when the system load is light.

Interestingly, some results are marked as {($\approx$)}, indicating that there is no statistically significant difference between the test performance across different utilisation levels. These cases typically occur between neighbouring utilisation levels (e.g., 0.75 and 0.85, or 0.95 and 0.99). This suggests that GP-evolved rules can maintain a certain degree of robustness when the workload conditions are similar, as the distributions of decision situations do not differ drastically. In such neighbouring settings, the shop floor states (e.g., queue lengths, idle rates) remain comparable, and the decision-making logic learned by GP under one utilisation level can still be effectively applied to another. Therefore, although the overall cross-utilisation generalisation is limited, GP shows local generalisation ability within similar workload ranges.

\subsubsection{Due Date Factor}
\label{due date}

From relaxed to tight schedules, five due date factors are set, specifically 1.0, 1.2, 1.5, 2.0, and 4.0, to investigate the generalisation ability of GP-evolved rules under different scheduling tightness levels.

\begin{table}[t] 
	\centering
	\footnotesize 
	\caption{The mean (standard deviation) of objective values on test instances according to 30 independent runs \textbf{across different due date factors}.}
	
	\vspace{-3mm}	
	\label{tab:duedate} 
	\resizebox{\textwidth}{!}{
		\newcommand{\tabincell}[2]{\begin{tabular}{@{}#1@{}}#2\end{tabular}}
		\begin{tabular} { cccccc} 
			\toprule 
			% --- 表头行（已加粗，使用 Due Date Factor） ---
			\bfseries \footnotesize 	
			\diagbox[width=4em]{Test}{Train} &
			\bfseries \footnotesize $\langle$1.0$\rangle$ &
			\bfseries \footnotesize $\langle$1.2$\rangle$ &	
			\bfseries \footnotesize $\langle$1.5$\rangle$	 &	
			\bfseries \footnotesize $\langle$2.0$\rangle$ &	
			\bfseries \footnotesize $\langle$4.0$\rangle$ \\
			\midrule
			
			% --- 第 1 行：Test <1.0> ---
			\bfseries $\langle$1.0$\rangle$ & 
			\colorbox{lightgray}{484.44(23.18)} & 
			481.88(18.92) {($\approx$)} & % (=) -> \approx
			498.07(32.01) {($\downarrow$)} & % (-) -> \downarrow
			512.49(31.17) {($\downarrow$)} & % (-) -> \downarrow
			771.72(168.44) {($\downarrow$)} \\ % (-) -> \downarrow
			
			% --- 第 2 行：Test <1.2> ---
			\bfseries $\langle$1.2$\rangle$ & 
			384.78(23.59) {($\approx$)} &	 % (=) -> \approx
			\colorbox{lightgray}{380.88(18.69)} & 
			396.23(31.68){($\approx$)}&	 % (-) -> \downarrow
			409.13(30.04){($\downarrow$)}&  % (-) -> \downarrow
			647.13(153.95) {($\downarrow$)} \\ % (-) -> \downarrow
			
			% --- 第 3 行：Test <1.5> ---
			\bfseries $\langle$1.5$\rangle$ &
			278.09(26.52) {($\approx$)} &	 % (=) -> \approx
			271.21(16.15){($\approx$)} &   % (=) -> \approx
			\colorbox{lightgray}{282.41(28.43)} &	
			289.74(26.38){($\downarrow$)} & % (-) -> \downarrow
			486.64(126.94) {($\downarrow$)} \\ % (-) -> \downarrow
			
			% --- 第 4 行：Test <2.0> ---
			\bfseries $\langle$2.0$\rangle$& 
			177.83(20.52){($\downarrow$)} &	 % (=) -> \approx
			171.04(9.57){($\approx$)} &   % (=) -> \approx
			174.40(18.19){($\approx$)}&	  % (=) -> \approx
			\colorbox{lightgray}{172.09(17.98)} & 
			293.52(86.32) {($\downarrow$)} \\ % (-) -> \downarrow
			
			% --- 第 5 行：Test <4.0> ---
			\bfseries $\langle$4.0$\rangle$& 
			56.97(4.77) {($\downarrow$)} &	 % (-) -> \downarrow
			55.56(4.47){($\downarrow$)} &   % (=) -> \approx
			51.62(3.32){($\downarrow$)}&	  % (+) -> \uparrow
			46.23(3.20){($\downarrow$)}&    % (+) -> \uparrow
			\colorbox{lightgray}{33.05(12.10)}\\
			
			\bottomrule 
		\end{tabular}
	}
\end{table}

\begin{table}[t] 
	\centering
	\footnotesize 
	\caption{The mean (standard deviation) of objective values on test instances according to 30 independent runs \textbf{across different batch sizes}.}
	
	\vspace{-3mm}	
	\label{tab:testBatchSize_NewData} 
	\resizebox{\textwidth}{!}{
		\newcommand{\tabincell}[2]{\begin{tabular}{@{}#1@{}}#2\end{tabular}}
		\begin{tabular} { cccccc} 
			\toprule 
			% --- 表头行（已加粗，使用 Batch Size） ---
			\bfseries \footnotesize 	
			\diagbox[width=4em]{Test}{Train} &
			\bfseries \footnotesize $\langle$Single$\rangle$ &
			\bfseries \footnotesize $\langle$Small$\rangle$ &	
			\bfseries \footnotesize $\langle$Medium$\rangle$ &	
			\bfseries \footnotesize $\langle$Large$\rangle$ &	
			\bfseries \footnotesize $\langle$Huge$\rangle$ \\
			\midrule
			
			% --- 第 1 行：Test <Single> ---
			\bfseries $\langle$Single$\rangle$ & 
			\colorbox{lightgray}{70.28(3.76)} & 
			74.07(4.23) {($\downarrow$)} &   % (-) -> \downarrow
			107.09(50.21) {($\downarrow$)} & % (-) -> \downarrow
			151.64(67.91) {($\downarrow$)} & % (-) -> \downarrow
			190.79(68.65) {($\downarrow$)} \\ % (-) -> \downarrow
			
			% --- 第 2 行：Test <Small> ---
			\bfseries $\langle$Small$\rangle$ & 
			471.56(64.48) {($\downarrow$)} &	 % (-) -> \downarrow
			\colorbox{lightgray}{380.88(18.69)} & 
			409.05(51.98){($\downarrow$)}&	  % (-) -> \downarrow
			448.43(59.72){($\downarrow$)}&   % (-) -> \downarrow
			486.72(54.34) {($\downarrow$)} \\ % (-) -> \downarrow
			
			% --- 第 3 行：Test <Medium> ---
			\bfseries $\langle$Medium$\rangle$ &
			1270.55(232.16) {($\downarrow$)} & % (-) -> \downarrow
			932.16(57.54){($\approx$)} &    % (+) -> \uparrow
			\colorbox{lightgray}{947.34(77.40)} &	
			979.89(80.33){($\downarrow$)} & % (-) -> \downarrow
			1010.34(65.80) {($\downarrow$)} \\ % (-) -> \downarrow
			
			% --- 第 4 行：Test <Large> ---
			\bfseries $\langle$Large$\rangle$& 
			1890.08(361.75){($\downarrow$)} &	% (-) -> \downarrow
			1361.08(90.70){($\approx$)} &    % (+) -> \uparrow
			1365.94(103.33){($\approx$)}&	  % (=) -> \approx
			\colorbox{lightgray}{1390.35(106.96)} & 
			1410.56(87.53) {($\downarrow$)} \\ % (-) -> \downarrow
			
			% --- 第 5 行：Test <Huge> ---
			\bfseries $\langle$Huge$\rangle$& 
			3781.97(825.24) {($\downarrow$)} &	% (-) -> \downarrow
			2588.89(178.25){($\downarrow$)} &    % (+) -> \uparrow
			2564.61(189.81){($\approx$)}&	  % (+) -> \uparrow
			2574.72(204.07){($\approx$)}&    % (=) -> \approx
			\colorbox{lightgray}{2561.57(165.46)}\\
			
			\bottomrule 

		\end{tabular}
	}
\end{table}

Table~\ref{tab:duedate} shows the test performance across different due date factors. The results indicate that GP-evolved rules generally perform best when the training and test due date factors match or very similar (1.0, 1.2, and 1.5). As the discrepancy between training and test factors increases, performance tends to deteriorate. For example, rules trained under tight due date factors (e.g., 1.0 or 1.2) yield much higher objective values when applied to loose due date environments (e.g., 2.0 or 4.0), because these rules prioritise urgent jobs and may be overly aggressive, resulting in poor performance when slack is abundant. Conversely, rules evolved under loose due date factors underperform in tight due date settings, as they do not sufficiently prioritise urgent jobs.

\subsubsection{Batch Size}
From small to large batches, five batch sizes are considered: Single (1), Small (1–9, uniformly distributed), Medium (10–20), Large (20–30), and Huge (30–40), to examine the generalisation ability of GP-evolved rules across different batch sizes.

Table~\ref{tab:testBatchSize_NewData} presents the test performance across different batch sizes. The results show that rules trained on a specific batch size perform best when applied to the same size. Rules evolved on small batches tend to underperform on larger batches due to increased resource contention and coordination complexity, whereas rules trained on large or huge batches perform poorly on small batches. The underlying reason lies in the change of decision point distributions across batch sizes. Larger batches cause higher instantaneous congestion, longer queues, and heavier machine workloads, creating decision situations more extreme than those in smaller batches. These shifts in decision point distributions explain the sensitivity of GP-evolved rules to batch size. Moreover, the performance differences between neighbouring batch sizes (e.g., Medium and Large) are relatively small, likely because their decision point distributions are similar.

\subsection{Different Parameter Distributions}

In this part, the training and test instances differ in the parameter distribution. Specifically, five types of distributions are considered: Exponential, Gamma, Lognormal, Normal, and Uniform. Table~\ref{tab:testDistribution_NewData} shows the test performance across different parameter distributions. The results show a clear pattern: GP-evolved rules achieve the best performance when tested on instances with the same distribution as they were trained on. When the training and test distributions differ, performance tends to deteriorate, indicating that the rules are highly sensitive to the underlying parameter distribution. This highlights that GP-evolved rules are effective at capturing patterns specific to a particular distribution but have limited generalisation across different distribution types.

\begin{table}[t] 
	\centering
	\footnotesize 
	\caption{The mean (standard deviation) of objective values on test instances according to 30 independent runs \textbf{across different distribution types}.}
	
	\vspace{-3mm}	
	\label{tab:testDistribution_NewData} 
	\resizebox{\textwidth}{!}{
		\newcommand{\tabincell}[2]{\begin{tabular}{@{}#1@{}}#2\end{tabular}}
		\begin{tabular} { cccccc} 
			\toprule 
			% --- 表头行（已加粗，使用分布类型） ---
			\bfseries \footnotesize 	
			\diagbox[width=4em]{Test}{Train} &
			\bfseries \footnotesize $\langle$Exponential$\rangle$ &
			\bfseries \footnotesize $\langle$Gamma$\rangle$ &	
			\bfseries \footnotesize $\langle$LogNormal$\rangle$	 &	
			\bfseries \footnotesize $\langle$Normal$\rangle$ &	
			\bfseries \footnotesize $\langle$Uniform$\rangle$ \\
			\midrule
			
			% --- 第 1 行：Test <E> ---
			\bfseries $\langle$Exponential$\rangle$ & 
			\colorbox{lightgray}{14.78(0.93)} & 
			18.69(3.22) {($\downarrow$)} &   % (-) -> \downarrow
			17.99(3.33) {($\downarrow$)} & % (-) -> \downarrow
			110.32(500.77) {($\downarrow$)} & % (-) -> \downarrow
			16.05(1.26) {($\downarrow$)} \\ % (-) -> \downarrow
			
			% --- 第 2 行：Test <G> ---
			\bfseries $\langle$Gamma$\rangle$ & 
			297.21(52.13) {($\downarrow$)} &	 % (-) -> \downarrow
			\colorbox{lightgray}{271.23(23.97)} & 
			282.96(24.13){($\downarrow$)}&	   % (=) -> \approx
			355.87(446.96){($\downarrow$)}&    % (=) -> \approx
			281.60(97.14) {($\downarrow$)} \\ % (-) -> \downarrow
			
			% --- 第 3 行：Test <L> ---
			\bfseries $\langle$LogNormal$\rangle$ &
			380.49(54.28) {($\downarrow$)} &	 % (-) -> \downarrow
			375.93(57.56){($\downarrow$)} &    % (=) -> \approx
			\colorbox{lightgray}{339.74(24.52)} &	
			345.55(22.46){($\downarrow$)} & % (-) -> \downarrow
			580.75(1378.60) {($\downarrow$)} \\ % (-) -> \downarrow
			
			% --- 第 4 行：Test <N> ---
			\bfseries $\langle$Normal$\rangle$& 
			207.82(26.68){($\downarrow$)} &	% (-) -> \downarrow
			206.34(29.97){($\downarrow$)} &    % (=) -> \approx
			189.05(12.34){($\downarrow$)}&	   % (+) -> \uparrow
			\colorbox{lightgray}{181.46(2.31)} & 
			198.99(94.56) {($\downarrow$)} \\ % (-) -> \downarrow
			
			% --- 第 5 行：Test <U> ---
			\bfseries $\langle$Uniform$\rangle$& 
			439.81(63.59) {($\downarrow$)} &	% (-) -> \downarrow
			424.67(53.28){($\downarrow$)} &    % (=) -> \approx
			398.22(28.23){($\downarrow$)}&	   % (+) -> \uparrow
			401.31(17.71){($\downarrow$)}&    % (=) -> \approx
			\colorbox{lightgray}{380.88(18.69)}\\
			
			\bottomrule 
		\end{tabular}
	}
\end{table}

\subsection{Number and Distribution of Decision Points}

To investigate whether the number and distribution of decision points in DFJSS instances, as discussed in the results section, are the underlying factors influencing the generalisation of GP-evolved rules, we conducted a detailed analysis. The number of decision points refers to how many scheduling decisions a instance generates. The distribution of decision points refers to the characteristics of these decision points, including the feature values associated with each decision, e.g., the number of candidate machines for an operation, the workload or the number of waiting operations on a machine, and other relevant state features.

Two representative cases were selected: the experiments in Section~\ref{job} (varying job numbers) and Section~\ref{utilisation level} (varying utilisation levels). The workload of each machine is a key feature in DFJSS problems~\cite{zhang2020FS}, as it reflects the dynamic state of the shop floor. Therefore, machine workload is used to characterise the state of a decision point. Specifically, we use a fixed rule pair to run the simulation and focus on sequencing decision point moments when a machine becomes idle and select an operation from its waiting queue. At each of these points, the current machine workload is recorded. To mitigate randomness from a single instance, sequencing points are collected from 10 independent instances for analysis. After obtaining the workload information, we plot the distributions and calculate the overlap ratios between DFJSS instances of different types to quantify the similarity of their decision point distributions. Specifically, the overlap ratio represents the degree of distributional overlap between two instances.

\begin{figure}[t]
	\centering 
	\includegraphics[width= 1\textwidth]{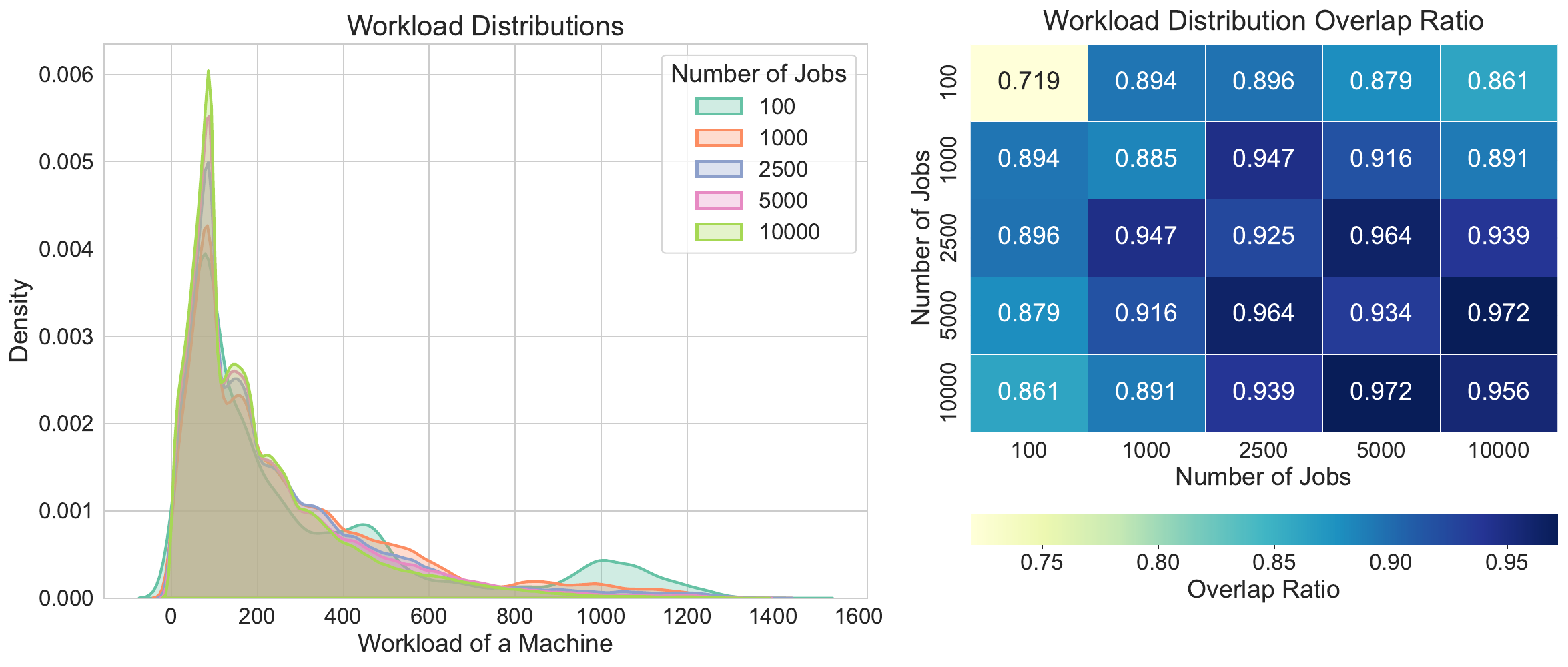}
	\vspace{-3mm}
	\caption{Workload distributions and overlap ratios among instances with different number of jobs.} \label{Fig. distributionJobs}
\end{figure}

Figure~\ref{Fig. distributionJobs} illustrates the workload distributions and overlap ratios among instances with different numbers of jobs. The distributions are highly similar, and the overlap ratios are quite high, indicating that the decision situations among these instances share a nearly identical structure. However, the number of decision points varies significantly with problem scale, the average numbers of decision points for the five job settings are 747, 6418, 23503, 31012, and 61673, respectively. This means that instances with more jobs provide substantially more decision samples for GP training, which helps explain why rules evolved on large-scale instances can generalise well to smaller problems. This observation is consistent with our expectation.

\begin{figure}[t]
	\centering 
	\includegraphics[width= 1\textwidth]{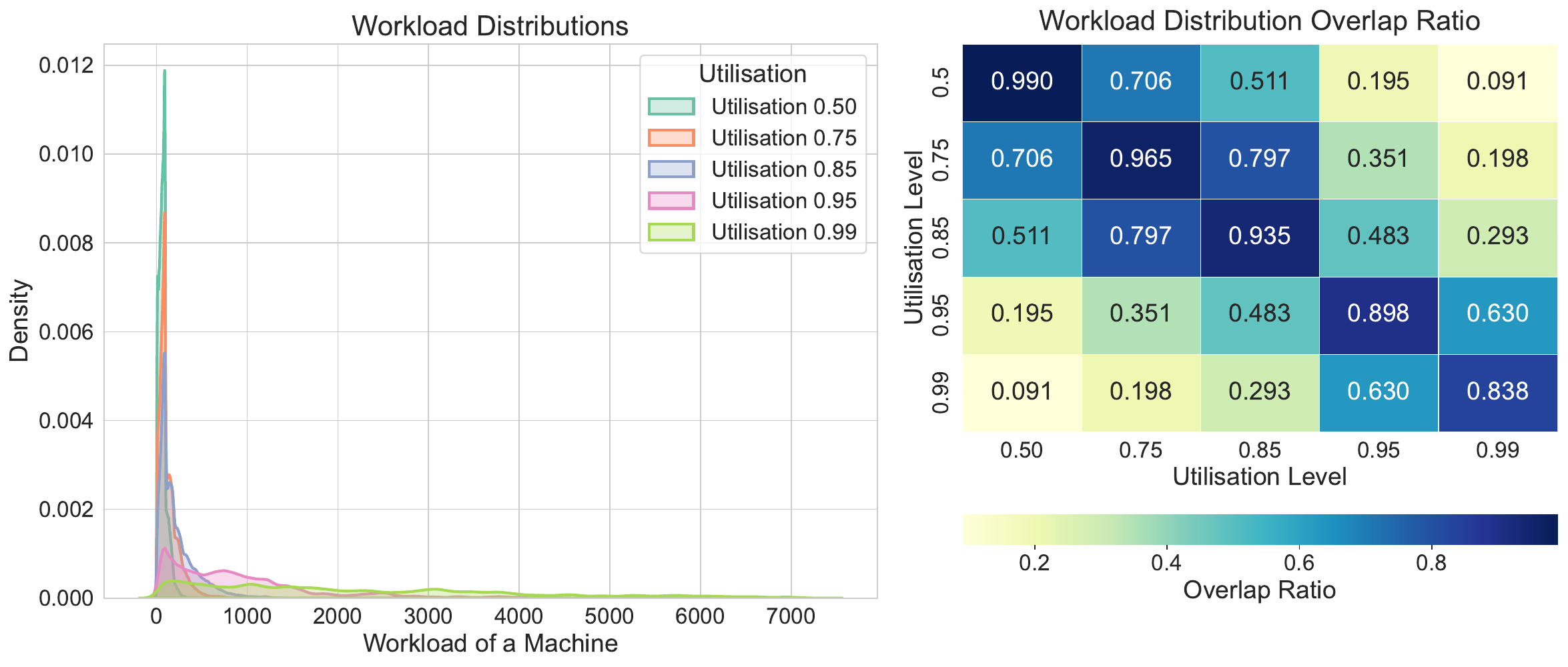}
	\vspace{-3mm}
	\caption{Workload distributions and overlap ratios among instances with different utilisation levels.} \label{Fig. distributionUtilisation}
\end{figure}

Figure~\ref{Fig. distributionUtilisation} illustrates the workload distributions and overlap ratios among DFJSS instances with different utilisation levels. As utilisation increases, the workload distributions become wider and shift toward higher values, reflecting greater resource contention and heavier system load. Under low utilisation (e.g., 0.50), workloads are concentrated near 100, indicating idle machines and low congestion. As utilisation level rises, the distributions spread and develop long tails, suggesting that some machines remain heavily loaded for extended periods.

The overlap ratio matrix quantifies the similarity between workload distributions. Diagonal values are close to 1.0, confirming that instances with the same utilisation level share nearly identical decision point distributions. In contrast, overlap ratios decline sharply as the difference between utilisation levels increases—for example, from 0.990 (0.50–0.50) to 0.091 (0.50–0.99). Neighbouring levels (e.g., 0.75–0.85 or 0.95–0.99) still exhibit relatively high overlaps (around 0.70–0.80), indicating similar shop-floor dynamics.

These findings are consistent with the performance trends in Table~\ref{tab:utilisation}. To further quantify this relationship, we calculated the Pearson and Spearman correlations between the overlap ratios and test performance for each utilisation level (each row). The Pearson coefficients are -0.879, -0.914, -0.850, -0.708, and -0.733, while the Spearman coefficients are -0.6, -0.9, -0.7, -0.6, and -0.9, respectively. These strong negative correlations indicate that a higher overlap between decision point distributions corresponds to better performance (i.e., lower objective values). In other words, the more similar the decision point distributions between training and test instances, the stronger the generalisation ability of GP-evolved rules. Overall, these results underscore the crucial role of distributional similarity in enhancing the generalisation of GP-evolved scheduling rules across DFJSS problems under varying conditions.

\section{Answers to the Proposed Questions}

According to the results presented in the previous section, the following summarises the answers to the research questions addressed in this paper.

\begin{itemize}
\item \textbf{How does \textit{problem scale} affect the generalisation ability of GP-evolved rules?}

When only the number of jobs changes, GP-evolved rules trained on large-scale instances generalise well to smaller ones. However, when the number of machines changes or when both jobs and machines vary, good generalisation is observed only between instances of the same or very similar scale.

\item \textbf{How does \textit{workshop parameter} influence the generalisation ability of GP-evolved rules?}

For all examined workshop parameters (utilisation level, due date factor, and batch size), the best performance occurs when the training and testing conditions are identical or similar. As the gap between them widens, performance generally declines.

\item \textbf{How does \textit{parameter distribution} impact the generalisation ability of GP-evolved rules?}

GP-evolved rules exhibit strong generalisation only when the training and testing instances share the same parameter distribution. When the underlying distribution differs, performance declines sharply.

\item \textbf{What underlying factors contribute to the observed performance differences, and what insights can be drawn from the results?}

Two key factors contribute to the observed generalisation behaviour:
(1) the number of decision points, and
(2) the distribution of decision points encountered during training and test. When the decision point distributions are similar, rules trained on instances with a larger number of decision points generalise well to smaller problems, as they are exposed to a broader range of decision contexts. However, when these distributions differ substantially, the number of decision points becomes less influential. In such cases, the dominant factor is the difference in distributional characteristics, which determines the structure of the scheduling environment and the types of decision-making situations that GP needs to handle.

These findings yield several practical and theoretical insights.

\begin{itemize}
\item When two scenarios produce highly similar decision-point distributions (e.g., utilisation levels of 0.95 and 0.99), rules trained in one scenario can generalise well to the other. This suggests that distributional similarity could serve as a predictor of the relationship between two instances, enabling us to estimate whether a rule will generalise from one instance to another without actually running the simulation.

\item Generalisation is most affected by instances with different parameter distributions or large gaps in key problem parameters, both of which substantially change the decision point distribution.

\item Current GP methods still struggle to generalise across instances with substantially different parameters, scales, or distributions, highlighting the need for approaches that evolve scheduling rules with stronger and more consistent generalisation ability.
\end{itemize}
\end{itemize}

\section{Conclusions and Future Work}

The goal of this paper was to investigate the generalisation ability of GP-evolved rules in DFJSS problems. This objective has been successfully achieved through a comprehensive set of experiments conducted on diverse DFJSS instances, which enabled a systematic analysis of performance differences and the identification of the underlying factors contributing to these results.

The findings reveal that GP-evolved rules can generalise well only when the training and testing instances share similar characteristics or distributions, such as having the same number of machines while varying the number of jobs, or similar utilisation levels, due date factors, and batch sizes. In addition, GP-evolved rules trained on large-scale instances can generalise well to smaller ones. However, when the distributions or structural properties differ significantly, the generalisation performance deteriorates markedly. Overall, GP-evolved rules demonstrate limited generalisation ability across different types of instances.

An interesting future research direction is to combine GP with lifelong learning techniques. In real-world applications, different instances appear over time, naturally leading to a lifelong learning setting; instead of training GP on a single type of instance, the system could continuously learn from DFJSS instances with varying scales and parameter distributions. This approach has the potential to evolve scheduling rules with stronger generalisation ability across heterogeneous environments.

% % References will then be sorted and formatted in the correct style.
\bibliographystyle{unsrt}
\bibliography{mybib.bib}
\end{document}